\documentclass[conference]{IEEEtran}
\IEEEoverridecommandlockouts

\usepackage{tikz}
\usepackage{textcomp}
\usepackage{hyperref}
\usepackage{lipsum}

\newcommand\copyrighttext{%
  \footnotesize \textcopyright 2022 IEEE. Personal use of this material is permitted.
  Permission from IEEE must be obtained for all other uses, in any current or future 
  media, including reprinting/republishing this material for advertising or promotional 
  purposes, creating new collective works, for resale or redistribution to servers or 
  lists, or reuse of any copyrighted component of this work in other works. 
  DOI: \href{https://ieeexplore.ieee.org/document/10069169}{10.1109/ICMLA55696.2022.00090}
  }
\newcommand\copyrightnotice{%
\begin{tikzpicture}[remember picture,overlay]
\node[anchor=south,yshift=10pt] at (current page.south) {\fbox{\parbox{\dimexpr\textwidth-\fboxsep-\fboxrule\relax}{\copyrighttext}}};
\end{tikzpicture}%
}

\usepackage{cite}
\usepackage{amsmath,amssymb,amsfonts}
\usepackage{algorithmic}
\usepackage{graphicx}
\usepackage{textcomp}
\usepackage{xcolor}
\usepackage{multirow}
\def\BibTeX{{\rm B\kern-.05em{\sc i\kern-.025em b}\kern-.08em
    T\kern-.1667em\lower.7ex\hbox{E}\kern-.125emX}}
\begin{document}

\title{On the Generalizability of ECG-based Stress Detection Models\\
\thanks{This work is partially funded by the European Union’s Horizon 2020 research and innovation programme under grant agreement No 847926 MindBot}
}

\author{\IEEEauthorblockN{Pooja Prajod}
\IEEEauthorblockA{\textit{Human-Centered Artificial Intelligence} \\
\textit{University of Augsburg}\\
Augsburg, Germany \\
pooja.prajod@uni-a.de}
\and
\IEEEauthorblockN{Elisabeth Andr\'e}
\IEEEauthorblockA{\textit{Human-Centered Artificial Intelligence} \\
\textit{University of Augsburg}\\
Augsburg, Germany \\
elisabeth.andre@uni-a.de}
}

\maketitle
\copyrightnotice

\begin{abstract}
Stress is prevalent in many aspects of everyday life including work, healthcare, and social interactions.
Many works have studied handcrafted features from various bio-signals that are indicators of stress.
Recently, deep learning models have also been proposed to detect stress.
Typically, stress models are trained and validated on the same dataset, often involving one stressful scenario.
However, it is not practical to collect stress data for every scenario.
So, it is crucial to study the generalizability of these models and determine to what extent they can be used in other scenarios.
In this paper, we explore the generalization capabilities of Electrocardiogram (ECG)-based deep learning models and models based on handcrafted ECG features, i.e.,  Heart Rate Variability (HRV) features. 
To this end, we train three HRV models and two deep learning models that use ECG signals as input.
We use ECG signals from two popular stress datasets - WESAD and SWELL-KW - differing in terms of stressors and recording devices.
First, we evaluate the models using leave-one-subject-out (LOSO) cross-validation using training and validation samples from the same dataset.
Next, we perform a cross-dataset validation of the models, that is, LOSO models trained on the WESAD dataset are validated using SWELL-KW samples and vice versa. 
While deep learning models achieve the best results on the same dataset, models based on HRV features considerably outperform them on data from a different dataset.
This trend is observed for all the models on both datasets.
Therefore, HRV models are a better choice for stress recognition in applications that are different from the dataset scenario.
To the best of our knowledge, this is the first work to compare the cross-dataset generalizability between ECG-based deep learning models and HRV models.

\end{abstract}


\section{Introduction}
Stress recognition research has become an important part of affective computing, especially in applications involving human-computer interaction~\cite{wesad}. 
Long-term stress has severe consequences and hence, there is a need for automatic stress recognition to detect stress early~\cite{wesad,annbetter}.
Stress stimuli or stressors trigger physiological responses in people which can be detected through different bio-signals such as Electrocardiogram (ECG) and Electrodermal Activity (EDA)~\cite{wesad, annbetter, swell}.
So, stress recognition research is further facilitated by the increasing popularity of wearable sensors that can unobtrusively collect real-time bio-signal data~\cite{swell2, wesadeval3}.

ECG is one of the most common bio-signal used in stress and affect recognition~\cite{ecgemotion, wesadeval3}.
There are two popular approaches to detect stress from ECG - models based on hand-crafted Heart Rate Variability (HRV) features~\cite{wesad, annbetter, swell2, hrv1} and deep learning models~\cite{deepecgnet, ecgemotion, wesadeval3}.
HRV features and their relationship with stress have been studied thoroughly~\cite{hrv3, hrv4}.
They have also been validated as indicators of stress in different stressful conditions~\cite{wesad, swell2, DeepMLcomp}.
However, cleaning the ECG signal and computing the HRV features often require specific domain knowledge~\cite{wesadeval3}.
This paved the way for deep learning models, which typically have convolution layers for automatic feature extraction.

We say an ECG-based stress recognition model has good generalization capability if it performs well on samples collected using different sensor devices under different stress conditions. 
It is crucial to evaluate the generalizability of a model as it is not possible to collect stress data and train specialized models in every scenario. 
In some cases, the models have to be trained on an available dataset and deployed in a scenario different from the training dataset.
For example, a neuro-rehabilitation use-case described in ~\cite{rhythmNeurorehab} employs an agent which adapts exercises by taking into account the stress level of the patient.
Due to ethical considerations, it is difficult to collect a dataset by stressing the patients during a rehabilitation session.
Another example to consider is stress recognition for special groups of people, like people with autism spectrum disorder (ASD), dementia, etc.
Often, there is a lack of stress datasets that includes data collected from these groups of people.
Moreover, there could be differences in the intensity or the characteristics of stress responses of the people belonging to these groups.
For instance, one of the datasets we consider in this study is the WESAD dataset~\cite{wesad}, which uses social evaluation as a stressor.
But, in~\cite{asdTSST}, the authors found that children with ASD had blunted physiological stress response to social evaluation stressor.

In this work, we investigate if the models trained on one stress dataset can detect stress in another dataset. 
Specifically, among ECG deep learning models and HRV models, we determine if one group outperforms the other in detecting stress samples from another scenario.

\section{Related Work}
Due to the health consequences of stress, there is extensive research on stress recognition.
It is beyond the scope of this work to summarize the numerous works that improve stress recognition.
So, we focus on works that compare various models or stress datasets to gain insights into trends pertaining to their performance.

There are multiple feature-based models proposed for stress recognition in various works.
Bobade and Vani~\cite{annbetter} compare the stress recognition performance of various machine learning models trained on hand-crafted features from various physiological signals. 
They use the WESAD dataset~\cite{wesad} to train K-Nearest Neighbour (KNN), Linear Discriminant Analysis (LDA), Random Forest Classifier (RFC), Support Vector Machine (SVM), etc. 
They also propose a simple feed-forward Artificial Neural Network (ANN) trained on the same input. 
Their comparison shows that ANN achieves higher accuracy than other models. 

As mentioned before, there are two main types of stress recognition models - deep neural networks and feature-based machine learning models. Naturally, questions arise on whether one type is better than the other.
Zhang et al.~\cite{DeepMLcomp} address this question by studying the performance of a deep neural network and feature-based models on a dataset they collected.
They propose a stress recognition model consisting of both convolutional neural networks (CNN) and bidirectional long short-term memory (BiLSTM). 
For comparison, they extract HRV features and train popular machine learning models like SVMs, RFC, Ada Boost, etc. 
The CNN-LSTM model takes $10\ s$ of raw ECG signal, whereas the other machine learning models use HRV features extracted from $60\ s$ of ECG data.
Zhang et al. demonstrate that deep neural networks significantly outperform HRV-based models.

Dzie{\.z}yc et al.~\cite{CNNbetterLSTM} compare various deep learning models on their performance in emotion recognition tasks (including stressful condition). 
An extensive study is performed on four different datasets, separately.
They chose an input signal length of $50-60\ s$, which is longer than the typical input length for deep learning models.  
They note that CNN-based models tend to perform better than LSTM-based models.

All the above works train and test the stress recognition models on the same dataset. 
Cho et al.~\cite{stressTL} consider two datasets differing in size and train ECG-based deep learning models to detect stress.
They propose a transfer learning approach, which involves training a model on the bigger dataset and then fine-tuning it on the smaller dataset. 
They observe that the stress recognition on the smaller dataset improves through transfer learning.
Other than the size, the datasets were very similar (e.g. same ECG sensor and configuration). 
The authors note that when data from other datasets are used, their model shows high bias to the type of stressor and a dependency on the sensor used.
In line with this observation, Liapis et al.~\cite{edaComp} demonstrate that a high stress recognition accuracy on one dataset does not necessarily translate to high accuracy in another dataset.
To this end, they extract Skin Conductance (SC) features from the WESAD dataset~\cite{wesad} and train four machine learning models for stress recognition.
These models achieve high accuracy while testing on the WESAD dataset.
However, they did not achieve good results on input signals from a different dataset (UX evaluation dataset).
Since the UX evaluation dataset is annotated primarily for emotion and not stress, it is difficult to conclude about the generalizability of the models.
Nevertheless, their observation highlights the need for cross-dataset evaluations and assessing the generalizability of the stress recognition models.

As a first step towards combining stress datasets for developing generic models, Baird et al.~\cite{speechComp} evaluate three datasets on their ability to predict cortisol values. 
Cortisol values are considered the ground truth for stress response.
As they note, the scales of cortisol values of the datasets are incompatible and thus, a cross-dataset evaluation is not feasible.
However, all three datasets were collected through similar Trier Social Stress Test (TSST) procedures.
So, the responses in each condition of the test are expected to be similar and therefore, the trends in predicted cortisol values can be compared.
To this end, they extract features from the speech signals in the datasets and train models for each dataset. 
They highlight the feasibility of using speech signals from one dataset as predictors of stress in another dataset.

\section{Approach}
Deep learning models trained directly on the ECG signal typically outperform hand-crafted HRV features on a given dataset~\cite{DeepMLcomp}. 
However, it remains unexplored if these deep learning models perform equally well in cross-dataset evaluations. 
To investigate this, we train $5$ stress recognition models - two deep learning models using ECG signals as input, and three models based on hand-crafted HRV features. 
First, we train and evaluate the stress models on the same dataset using leave-one-subject-out (LOSO) cross-validation. 
We perform this evaluation on two different datasets.
Then, we evaluate the LOSO models trained on dataset A using samples from the other dataset B (cross-dataset evaluation) to assess their generalization capabilities. 
Baird et al.~\cite{speechComp} note that machine learning models can benefit from combining stress datasets as it increases the data available for training.
It has not been investigated if this holds true if the datasets are vastly different, especially in terms of the stressors, the intensity of stress experienced, and the brand of sensors used.
So additionally, we train the models on a combined dataset (merging samples from the two datasets) and evaluate them using LOSO validation.

\subsection{Datasets}
\subsubsection{WESAD}
WESAD~\cite{wesad} is a multimodal dataset that contains motion (ACC) and physiological (ECG, EDA, etc.) signals, which were collected using chest-worn RespiBan and wrist-worn Empatica E4 devices. 
The data was collected from $15$ participants under three conditions: baseline, stress, and amusement. 
Stress was elicited using the Trier Social Stress Test (TSST) involving public speaking and mental arithmetic (counting down from $2023$ by steps of $17$) tasks. 
The stress condition lasted for about $10$ minutes. The amusement condition was around $6.5$ minutes long, where the participants watched funny video clips. 
In this work, we use the ECG data from the chest-worn device, sampled at $700\ Hz$. 
To be consistent with the labels used by the authors, we consider baseline and amusement conditions as the no-stress class.

\subsubsection{SWELL-KW} 
SWELL knowledge work dataset~\cite{swell} contains data collected from $25$ participants who did typical office tasks (writing reports and making a presentation) under three conditions - neutral, email interruptions, time pressure. 
During the email interruption session, $8$ emails were sent - many were irrelevant, and some required a reply. 
In the time pressure session, the participants had to complete their tasks in $2/3$rd of the time allotted for the neutral session. 
The neutral and email interruption sessions lasted for around $45$ minutes each, whereas the time pressure session lasted for around $30$ minutes. 
We use the ECG signals (sampled at $2048\ Hz$), which were collected using a TMSI Mobi device. The participants did not report feeling stressed in any of the conditions. 
However, they indicated higher temporal demand (they felt time pressure due to the pace of the task) during the time pressure session. 
In a subsequent study~\cite{swell2}, the authors labelled the data from email interruptions and time pressure sessions as stress and the neutral session as no-stress for a binary stress classification task. 
Hence, we also consider the data belonging to email interruptions and time pressure sessions as stress samples.

\subsection{Classification Models}
We describe our stress detection models and their training parameters below:

\subsubsection{Deep ECGNet} 
This is a CNN-LSTM stress detection model proposed in~\cite{deepecgnet}. 
The idea of CNN-LSTM networks is to use the CNN layers as feature extractors and train the LSTM layers to learn temporal patterns in the extracted features. 
The model has an initial convolution block containing a 1D convolutional layer, a pooling layer, a dropout layer, and a batch normalization layer. 
The activation function for the convolution layer is rectified linear unit (ReLU). 
The 1D convolution layer has $50$ filters with a kernel size corresponding to $0.6\ s$. 
The pooling layer has a size equivalent to $0.8\ s$ of data. 
For a $256\ Hz$ input, kernel size is $154$, and pooling size is $205$. 
The convolution block is followed by two LSTM layers and a final prediction layer. 
The first LSTM layer has $32$ units and the second one has $16$ units. 
We add a dropout layer and a batch normalization layer between the two LSTM layers. 
The activation function for the LSTM layers is Tanh, and for the prediction layer is Softmax. 
We use a dropout rate of $0.2$ for both dropout layers.

\subsubsection{ECG Emotion Recognition Model}
This CNN model is proposed in~\cite{ecgemotion} for emotion recognition on various datasets, including WESAD and SWELL-KW. 
The model has three convolution blocks, each block consisting of two 1D convolution layers and a pooling layer. 
The convolution layers belonging to a block have identical parameters such as kernel size and the number of filters. 
From block 1 to block 3, the number of filters are $32$, $64$, and $128$, whereas the kernel sizes are $32$, $16$, and $8$. 
The pooling layers are of size $8$ with strides of $2$.
The convolution blocks are followed by two fully connected layers with $128$ nodes each. 
We add a dropout layer (dropout rate $=\ 0.6$) after each fully connected layer. 
Finally, the model is connected to a stress prediction layer with Softmax activation.
All the convolution layers and the fully connected layers have ReLU activation. 

\subsubsection{Multi-Layer Perceptron} 
This is a simple neural network with an input layer, two hidden layers, and a prediction layer. 
The hidden layers have $12$ and $6$ nodes. 
The activation function for hidden layers is ReLU and for the prediction layer is Sigmoid.
To prevent over-fitting, we add a dropout layer (dropout rate $=\ 0.2$) after the input layer. 

\subsubsection{RFC}
This is an ensemble classifier that trains a certain number of decision trees on various subsets of the training set and uses their output to make a final prediction. 
This reduces over-fitting and improves the overall performance, even if individual classifiers are weak.
Similar to~\cite{wesad}, the number of decision trees (or estimators) is set to $100$ and the minimum number of samples for splitting a node is set to $20$.

\subsubsection{SVM}
This is a commonly used supervised learning model.
Similar to~\cite{swell2}, we use SVM with Radial basis function (Rbf) kernel.

We use Tensorflow to train neural networks and Scikit-learn to train other machine learning models. 
For all the models, we use a weighted loss to tackle class imbalance in the training dataset.
For the neural network models, we use the Adadelta optimizer (learning rate $=\ 1.0$) and cross-entropy loss.
We train them for $200$ epochs with a batch size of $128$.

\subsection{Evaluation metrics}
We use \textit{F1-score} and \textit{Accuracy} metrics for evaluation. 
Accuracy is the ratio of number of correctly predicted samples to total number of samples in the test set. 
F1-score is computed as the harmonic mean of precision and recall. 
Precision is the number of correctly predicted samples of a class out of all the samples predicted to belong to the class.
Recall is the number of correctly predicted samples of a class out of all the samples belonging to the class.
To tackle the class imbalances in the datasets, we compute macro f1-score, i.e., compute f1-score for each class and average them. 
We perform within-dataset LOSO evaluation as it determines the generalizability of a model on data from unseen users.
However, this is not an indicator of generalizability of the models on data collected using a different sensor or a different stressor.
Hence, we evaluate the models using cross-dataset validation.
This validation involves training a model on a dataset A and evaluating it using samples from another dataset B.

\subsection{Pre-processing}
The data collected in the two datasets have different sampling rates. 
This is not a concern for HRV features, but the ECG-based deep learning models require the input lengths to be the same. 
To keep the data consistent for all models, we down-sample the ECG signals in both datasets to $256\ Hz$. 

There are various sources of noise in an ECG signal, including baseline wander, powerline interference, and EMG noise~\cite{noise, noise2}. 
Baseline wander is a low-frequency noise ($0.5 - 0.6\ Hz$) that causes the signal to drift up and down. 
It is typically removed using a high-pass filter~\cite{noise, noise2, ecgemotion}. 
Powerline interference is caused by the electromagnetic interference of the power source of the sensor device. 
A common technique to remove this noise is using a band-stop or notch filter with a notch frequency of $50$ or $60\ Hz$ (depending on the device)~\cite{noise, noise2}. 
EMG noise is a high-frequency noise due to muscle contractions and the subject's movement. 
This noise can be reduced by using moving average~\cite{noise}.

\begin{figure}[htbp]
\centerline{\includegraphics[width=0.29\textwidth]{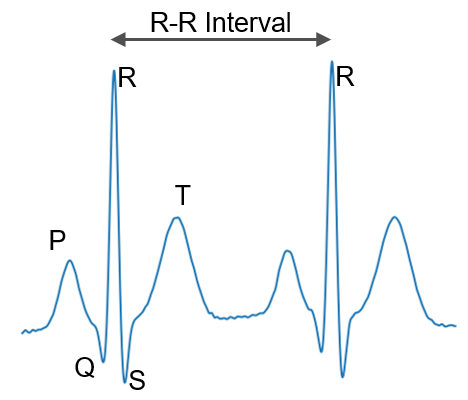}}
\caption{An example ECG signal with P-wave, QRS complex, and T-wave.}
\label{fig:ecg_example}
\end{figure}

As illustrated in Figure~\ref{fig:ecg_example}, a beat of ECG signal consists of P-wave, QRS complex, and T-wave.
For stress recognition, we are mostly interested in the QRS complex. 
Elgendi et al.~\cite{qrsband} propose a frequency band of $8 - 20\ Hz$ for the best signal-to-noise ratio on QRS components. 
We apply a second-order Butterworth band-pass filter with the proposed frequency band. 
We note that this filter removes most of the noises described above as their frequencies are outside the chosen band. 

The next steps in pre-processing are choosing input lengths and normalization. 
These steps differ depending on whether we use the filtered ECG signal as input or perform HRV feature extraction. 
Many studies have demonstrated that deep learning stress models can achieve good performance even on ultra-short-term ECG signals~\cite{DeepMLcomp, stressTL, deepecgnet, ecgemotion}. 
Deep ECGNet and ECG Emotion models were designed and validated on $10\ s$ segments of ECG signals~\cite{deepecgnet, ecgemotion}. 
On the other hand, studies typically use $60\ s$ of ECG data to extract reliable HRV features~\cite{hrv1, wesad, hrv3}.
We use $10\ s$ long filtered ECG data (without overlap) as input to the deep learning models. 
For HRV feature extraction, we use $60\ s$ windows of data with $50\ s$ overlap. 
We use this overlap to balance the number of training samples available for all the models.

Since the ECG devices used in the two datasets are different, the values would be recorded on different scales. 
Moreover, the individual stress responses could be different for different participants~\cite{ecgemotion}.
To circumvent these issues, we perform a user-specific Min-Max normalization. 
However, normalization will not eliminate the impact of using different sensor devices for recording ECG.
For deep learning models, we perform normalization on the filtered ECG data. 
Whereas, for HRV features, we perform normalization for every feature. 
In real-time stress recognition applications, the entire data would not be available for normalization. 
Hence, we adapt the approach from~\cite{norm} and use $5$ minutes of baseline data to compute the normalization parameters (i.e. min and max values).

\subsection{HRV features}
To calculate HRV, we first have to find the peaks in the ECG signal. 
We use the algorithm proposed in~\cite{qrsband} for finding the peaks, i.e., maximum amplitude in the QRS complex (see Figure~\ref{fig:ecg_example}). 
The algorithm utilizes the knowledge that for healthy adults (1) a 
beat has only one QRS complex and (2) the duration of QRS is $80 - 120\ ms$. 
We compute the interval between successive R-R peaks of an ECG signal to obtain the HRV signal.
We calculate a total of $22$ known HRV features from the time domain, frequency domain, and poincaré plots~\cite{wesad, hrv1, hrv2, hrv3, hrv4}.
These features are listed in Table~\ref{tab:hrv_feats} along with their descriptions.
We use NeuroKit2~\cite{nk2} python library for computing these features.

\begin{table}[htbp]
\caption{List of extracted HRV features}
\begin{center}
\begin{tabular}{|p{1.25cm}|p{6.25cm}|}
\hline
\textbf{Feature} & \textbf{Description} \\
\hline
HR & Number of R peaks in $1$ minute\\
MeanNN & Mean of R-R intervals \\
MedianNN & Median of R-R intervals \\
MadNN & Median Absolute Deviation of R-R intervals \\
StdNN & Standard deviation of R-R intervals \\
CVNN & Ratio of StdNN to MeanNN \\
IQRNN & Inter-Quartile Range of R-R intervals \\
RMSSD & Root Mean Square of successive differences of R-R intervals \\
StdSD & Standard deviation of successive differences of R-R intervals \\
pNN50 & $\%$ of successive differences of R-R intervals $> 50\ ms$  \\
pNN20 & $\%$ of successive differences of R-R intervals $> 20\ ms$  \\
TINN & Triangular Interpolation of R-R intervals histogram \\
HTI & HRV Triangular Index  \\
\hline
LF & Power of low frequency band ($0.04\ Hz - 0.15\ Hz$) in HRV spectrum \\
HF & Power of high frequency band ($0.15\ Hz - 0.4\ Hz$) in HRV spectrum \\
LF/HF & Ratio of LF to HF power \\
LFn & Normalized low frequency power, LF/total power \\
HFn & Normalized high frequency power, HF/total power \\
\hline
SD1 & Spread of HRV on Poincaré plot perpendicular to the identity line \\
SD2 & Spread of HRV on Poincaré plot along the identity line \\
SD1/SD2 & Ratio of SD1 to SD2 \\
S & Area of ellipse formed in the HRV Poincaré plot \\
\hline
\end{tabular}
\label{tab:hrv_feats}
\end{center}
\end{table}

\section{Results and Discussions}
In this section, we present the results of our evaluations. 
First, we evaluate the models using LOSO validation. 
We also compare their performance with other works on the same dataset. 
The results of LOSO evaluation on WESAD and SWELL-KW datasets are tabulated in Tables~\ref{tab:wesad_loso} and~\ref{tab:swell_loso}, respectively.

\begin{table}[htbp]
\caption{LOSO evaluation of ECG-based stress models on WESAD dataset}
\begin{center}
\begin{tabular}{|c|c|c|}
\hline
\textbf{Model} & \textbf{F1-score} & \textbf{Accuracy} \\
\hline
LDA~\cite{wesad} & 0.813 & 0.854 \\
LDA~\cite{wesadeval1} & - & 0.887 \\
CNN (Spectrogram)~\cite{wesadeval2} & 0.794 & 0.824 \\
Transformer~\cite{wesadeval3} (without fine-tuning) & 0.697 & 0.804 \\
\hline
Our RFC & 0.813 & 0.863 \\
Our SVM & 0.832 & 0.871 \\
Our MLP & \textbf{0.859} & 0.895 \\
Our ECG Emotion model & 0.858 & 0.897 \\
Our Deep ECGNet & 0.857 & \textbf{0.908} \\
\hline
\end{tabular}
\label{tab:wesad_loso}
\end{center}
\end{table}

\begin{table}[htbp]
\caption{LOSO evaluation of ECG-based stress models on SWELL-KW dataset}
\begin{center}
\begin{tabular}{|c|c|c|}
\hline
\textbf{Model} & \textbf{F1-score} & \textbf{Accuracy} \\
\hline
SVM~\cite{swell2} & - & 0.589 \\
Transformer~\cite{wesadeval3} (without fine-tuning) & 0.588 & 0.581 \\
\hline
Our RFC & 0.644 & 0.670 \\
Our SVM & 0.609 & 0.639 \\
Our MLP & 0.668 & 0.689 \\
Our ECG Emotion model & 0.627 & 0.709 \\
Our Deep ECGNet & \textbf{0.688} & \textbf{0.755} \\
\hline
\end{tabular}
\label{tab:swell_loso}
\end{center}
\end{table}

As we expected, the deep learning models perform better than the other models in within-dataset evaluations. 
On the WESAD dataset, the performance difference is relatively small ($< 5\%$), whereas it is higher on the SWELL-KW dataset.
The authors of~\cite{DeepMLcomp} had a similar observation between machine learning models and a CNN-LSTM model, both trained on a stress dataset they acquired.
We also note that, among the models based on HRV features, the MLP model performs the best.
This is in line with~\cite{annbetter}, where a simple feed-forward network is shown to perform better than machine learning methods (e.g., SVM, RFC) in a multimodal stress recognition task.
Considering both F1-score and Accuracy, Deep ECGNet has the overall best performance in both WESAD and SWELL-KW within-dataset LOSO evaluations. 

Next, we evaluate our models using cross-database validation. 
That is, models trained on the WESAD dataset are tested on SWELL-KW data and vice versa.
The results of cross-dataset evaluations are presented in Tables~\ref{tab:wesad_cross} and~\ref{tab:swell_cross}.

\begin{table}[htbp]
\caption{Cross-dataset evaluation of WESAD models on SWELL-KW dataset}
\begin{center}
\begin{tabular}{|c|c|c|}
\hline
\textbf{Model} & \textbf{F1-score} & \textbf{Accuracy} \\
\hline
Our RFC & 0.467 & 0.483 \\
Our SVM & \textbf{0.535} & \textbf{0.538} \\
Our MLP & 0.478 & 0.49 \\
Our ECG Emotion model & 0.395 & 0.411 \\
Our Deep ECGNet & 0.391 & 0.418 \\
\hline
\end{tabular}
\label{tab:wesad_cross}
\end{center}
\end{table}

\begin{table}[htbp]
\caption{Cross-dataset evaluation of SWELL-KW models on WESAD dataset}
\begin{center}
\begin{tabular}{|c|c|c|}
\hline
\textbf{Model} & \textbf{F1-score} & \textbf{Accuracy} \\
\hline
Our RFC & \textbf{0.581} & 0.637 \\
Our SVM & 0.509 & \textbf{0.647} \\
Our MLP & 0.49 & 0.621 \\
Our ECG Emotion model & 0.342 & 0.385 \\
Our Deep ECGNet & 0.392 & 0.415 \\
\hline
\end{tabular}
\label{tab:swell_cross}
\end{center}
\end{table}

The results of cross-dataset evaluations are the opposite of within-dataset evaluations. 
The deep learning models perform much worse than models trained on the HRV features.
In cross-dataset validation of WESAD models, SVM achieves the best F1-score and Accuracy.
Among SWELL-KW models, the RFC model has the overall best performance in cross-dataset evaluation, considering both F1-score and Accuracy.

From Tables~\ref{tab:wesad_cross} and~\ref{tab:swell_cross}, it is clear that HRV-based models outperform deep learning models in predicting stress from a different dataset.
This could be attributed to deep learning models learning dataset-specific features and not generic stress features.
We note that stressors in the two datasets are different and thus, the stress responses may be different.
Additionally, the sensors used for collecting ECG data are also different.
All these factors could influence the low generalization capabilities of the deep learning models.
More focused studies and datasets are required to improve the generalizability of the deep learning models.
On the other hand, HRV features have been studied thoroughly and validated as indicators of stress across multiple datasets with different stressors.
Moreover, HRV is computed based on the QRS peak position and thus, is not influenced by the difference in sensors.

Based on our observations, we suggest employing HRV models when the application scenario is different from the dataset.
The deep learning models perform better than HRV models on both WESAD and SWELL-KW within-dataset evaluations. 
So, deep learning models are preferred when the input to the model is similar to its training data. 

Finally, we investigate if combining the stress datasets lead to better stress recognition. 
Combining the WESAD and SWELL-KW datasets results in ECG data of 37 participants.
We train and evaluate our models using the data from the combined dataset using LOSO validation.

\begin{table*}[htbp]
\caption{LOSO evaluation of models on combined WESAD and SWELL-KW datasets}
\begin{center}
\begin{tabular}{|c|c|c|c|c|c|c|}
\hline
\multirow{2}{*}{\textbf{Model}} & \multicolumn{2}{c|}{\textbf{WESAD subjects}} & \multicolumn{2}{c|}{\textbf{SWELL-KW subjects}} & \multicolumn{2}{c|}{\textbf{All subjects}} \\
\cline{2-7}
& \textbf{F1-score} & \textbf{Accuracy} & \textbf{F1-score} & \textbf{Accuracy} & \textbf{F1-score} & \textbf{Accuracy} \\
\hline
Our RFC & 0.758 & 0.793 & 0.647 & 0.671 & 0.692 & 0.720 \\
Our SVM & 0.732 & 0.796 & 0.605 & 0.633 & 0.657 & 0.699 \\
Our MLP & \textbf{0.758} & \textbf{0.813} & 0.657 & 0.677 & \textbf{0.698} & \textbf{0.732} \\
Our ECG Emotion model & 0.609 & 0.677 & 0.593 & 0.683 & 0.599 & 0.681 \\
Our Deep ECGNet & 0.692 & 0.711 & \textbf{0.695} & \textbf{0.739} & 0.694 & 0.728 \\
\hline
\end{tabular}
\label{tab:combined}
\end{center}
\end{table*}

The results of LOSO validation on the combined dataset is shown are Table~\ref{tab:combined}. 
The F1-score and Accuracy of every model are significantly worse than the corresponding WESAD models (see Table~\ref{tab:wesad_loso}).
All the models achieve slightly lower F1-score and Accuracy than the corresponding SWELL models (see Table~\ref{tab:swell_loso}).
Despite the increase in training data, combining these two datasets does not improve the individual dataset or overall stress recognition.
So, combining the datasets is not beneficial for either of the datasets; even detrimental in the case of the WESAD dataset.

\section{Conclusion}
Due to the health benefits of detecting and mitigating stress early, there is a need for accurate and robust stress recognition models.
The stressor and intensity of stress experienced by people are different for different stressful conditions.
This coupled with the ethical challenges of collecting stress data, especially for special groups like people with autism, escalates the need for stress models with good generalization capabilities. 
Using two publicly available stress datasets (WESAD and SWELL-KW), we assessed the generalizability of five stress recognition models - two ECG-based deep learning models and three HRV feature-based models.
We first evaluated the models using within-dataset LOSO validation, followed by a cross-dataset evaluation. 
We found that ECG-based deep learning models outperform the HRV-based models on both stress datasets. 
However, HRV-based models were significantly better at recognizing stress in cross-dataset evaluations.
So, the HRV-based stress recognition models seem to be the better option when the model is deployed in a scenario that is considerably different than the training data acquisition.
We also investigate if the stress recognition improves when the models are trained on a combined dataset. 
The stress recognition on SWELL-KW subjects did not improve by combining datasets. 
On the other hand, this led to significantly lower performance of all five models on the WESAD dataset.

The datasets we considered in this paper differ in many aspects including the type of stressor, the intensity of stress experienced, and the brand of the ECG sensor. 
In the future, we plan to extend our work by considering more datasets and comparing them by controlling some of the aspects (e.g. sensor device). This will help us gain insights into the impact of specific factors on the generalizability of stress models.

\bibliographystyle{ieeetr}
\bibliography{references}

\end{document}